\documentclass[runningheads]{llncs}

\usepackage{eccv}

\usepackage{eccvabbrv}
\usepackage{upgreek}
\usepackage{pifont}
\usepackage{caption}
\usepackage{epsfig}
\usepackage{graphicx}
\usepackage{booktabs}
\usepackage{multirow}
\usepackage{colortbl}
\usepackage{wrapfig}

\definecolor{tabtitle}{gray}{.8}
\definecolor{mygray}{RGB}{244, 251, 250}
\definecolor{rouse}{rgb}{0.981,0.961,0.941}

\usepackage[accsupp]{axessibility}  

\usepackage{hyperref}
\usepackage{orcidlink}

\begin{document}
\definecolor{convcolor}{HTML}{412F8A}
	\definecolor{resnetcolor}{HTML}{8DA0CB}
	\definecolor{vitcolor}{HTML}{fc8e62}
\newcommand{\convcolor}[1]{\textcolor{convcolor}{#1}}
	\newcommand{\vitcolor}[1]{\textcolor{vitcolor}{#1}}
	\newcommand{\cnn}{ConvNeXt}

	\newcommand{\vb}{\vitcolor{$\mathbf{\circ}$\,}}
    \newcommand{\cb}{\convcolor{$\bullet$\,}}
    \newcommand{\gr}{\rowcolor[gray]{.95}}
\title{Learning Camouflaged Object Detection from Noisy Pseudo Label} 

\titlerunning{Learning Camouflaged Object Detection from Noisy Pseudo Label}

\author{Jin Zhang\inst{1}\orcidlink{0009-0002-9662-7635} \and
Ruiheng Zhang\inst{1}\thanks{Corresponding author. \href{mailto:ruiheng.zhang@bit.edu.cn}{ruiheng.zhang@bit.edu.cn}}\orcidlink{0000-0002-5460-7196} \and
Yanjiao Shi\inst{2}\orcidlink{0000-0001-9689-4165} \and
Zhe Cao\inst{1}\orcidlink{0009-0001-8503-3041} \and
Nian Liu\inst{3}\orcidlink{0000-0002-0825-6081} \and 
Fahad Shahbaz Khan\inst{3,4}\orcidlink{0000-0002-4263-3143}
}

\authorrunning{Zhang Jin et al.}

\institute{Beijing Institute of Technology, Beijing, China \and
Shanghai Institute of Technology, Shanghai, China \and
Mohamed bin Zayed University of Artificial Intelligence, Abu Dhabi, UAE \and 
Linköping University, Linköping, Sweden
\\
\url{https://github.com/zhangjinCV/Noisy-COD}}

\maketitle

\begin{abstract}
Existing Camouflaged Object Detection (COD) methods rely heavily on large-scale pixel-annotated training sets, which are both time-consuming and labor-intensive. Although weakly supervised methods offer higher annotation efficiency, their performance is far behind due to the unclear visual demarcations between foreground and background in camouflaged images. In this paper, we explore the potential of using boxes as prompts in camouflaged scenes and introduce the first weakly semi-supervised COD method, aiming for budget-efficient and high-precision camouflaged object segmentation with an extremely limited number of fully labeled images. Critically, learning from such limited set inevitably generates pseudo labels with serious noisy pixels. To address this, we propose a noise correction loss that facilitates the model's learning of correct pixels in the early learning stage, and corrects the error risk gradients dominated by noisy pixels in the memorization stage, ultimately achieving accurate segmentation of camouflaged objects from noisy labels. When using only 20\% of fully labeled data, our method shows superior performance over the state-of-the-art methods.
\end{abstract}

\section{Introduction}
Camouflaged Object Detection (COD) aims to detect and segment objects that blend seamlessly into their environments, presenting a significant challenge due to the need to counter sophisticated camouflage tactics and distinguish subtle differences between objects and their surroundings. Recent advances in COD \cite{fan2020camouflaged, fan2020pranet, pang2022zoom, he2023camouflaged} have been driven by the availability of abundant segmentation labels. However, the labeling process for camouflaged objects is extremely labor-intensive, requiring about 60 minutes per image \cite{fan2020camouflaged}, which poses a major obstacle to this field's development. This challenge has led to a growing trend towards exploring Weakly Supervised COD (WSCOD) methods \cite{he2023weakly, liang2022tree, chibane2022box2mask}, utilizing simpler annotations such as points \cite{liang2022tree}, scribbles \cite{he2023weakly}, and boxes \cite{lee2021bbam}, to potentially reduce labeling costs. Despite these efforts, the high similarity between the foreground and background in camouflaged images means that these methods still lag far behind Fully Supervised COD (FSCOD) methods in performance.

\begin{figure}[t]
            \centering
            \includegraphics[width=1\columnwidth]{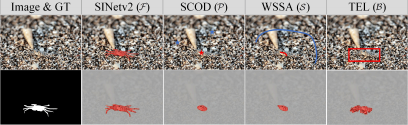}
            \caption{\textbf{Supervision behaviors and outputs}: $\mathcal{F}$, $\mathcal{P}$, $\mathcal{S}$, and $\mathcal{B}$ denote training with full, point, scribble, and box annotations. GT is the ground truth. The first row shows labels used for each method on the training image, with \textcolor{red}{red} for foreground and \textcolor{blue}{blue} for background (if needed). The second row presents outputs trained with these labels. }
		\label{fig:one}
        \end{figure}

Fig. \ref{fig:one} shows our comparative analysis of different supervision methods, which shows the classic challenges for COD tasks in the camouflaged image, such as high intrinsic similarity and unclear visual demarcations. These challenges severely impact model predictions; without accurate annotation as used in the fully supervised method, weakly supervised methods tend to produce serious false positive and negative predictions. Specifically, our findings indicate that sparse annotations, such as points and scribbles, hinder the discriminator's ability to accurately distinguish camouflaged objects from their environment, leading to a higher incidence of false negatives. Conversely, denser annotations like boxes often result in clearly false positives in scenarios with unclear visual demarcations between foreground and background.

Motivated by the above annotation properties, we explore the potential of utilizing box annotations as prompts for camouflaged object segmentation. Unlike points and scribbles, box annotations offer rich object information and are as cost-effective to provide as image-level and point-level annotations. We posit that using boxes as prompts offers reliability by \textbf{1}) masking complex backgrounds and reducing the level of camouflage, and \textbf{2}) indicating the approximate location of the object, thereby simplifying the model's search process for camouflaged objects. Thus, we formulate a new practical training scheme, Weakly Semi-Supervised Camouflaged Object Detection (WSSCOD), with box as prompt. In the WSSCOD task, we aim to achieve budget-efficient and high-performance camouflaged object segmentation using a small amount, such as 1\% number of total training set, of pixel-level annotations and corresponding box prompts.

\begin{figure}[t]
            \centering
            \includegraphics[width=1.0\columnwidth]{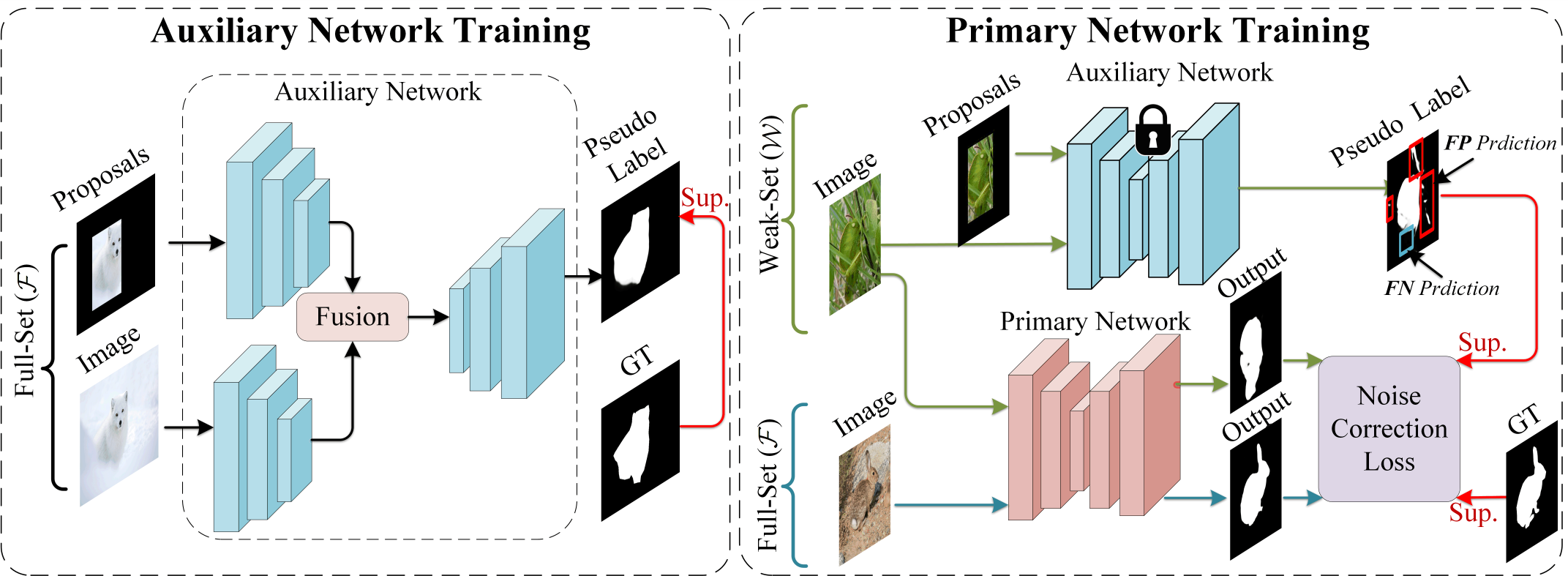}
            \caption{\textbf{Overview of the proposed WSSCOD}. Left part: Train the auxiliary network with full and box annotations. Right part: Use images and proposals as input to generate pseudo labels through the auxiliary network, where \textbf{\textit{FP}} (False Positive) and \textbf{\textit{FN}} (False Negative) predictions represent noisy pixels. Then, under the supervision of $\mathcal{L}_{NC}$, the primary network is trained using pixel-level annotations and pseudo labels.}
		\label{fig:two}
        \end{figure}

WSSCOD, as depicted in Fig. \ref{fig:two}, utilizes boxes as prompts to mask complex, similar backgrounds, delineating proposals for camouflaged objects. This method distinguishes camouflaged objects from their surroundings by focusing on the proposals, enabling the model to concentrate on the fine segmentation of object details rather than spending extra time searching for camouflaged objects first. Following this, we merge these proposals with the complete image to create complementary branches. This strategy reduces the impact of imprecise box locations on the model's decision boundaries. Ultimately, under the supervision of pixel-level annotations, a proposed COD model is trained with the complementary information, enabling it to generate high-quality pseudo labels with clear details for the remaining 99\% of the images.

Meanwhile, on an extremely limited amount (such as 1\%) of fully labeled data, the network often fails to represent the overall data distribution, resulting in rough and noisy pseudo labels. Moreover, when training with such noisy labels, we observe a distinct phenomenon: Initially, in the \textit{`early learning phase'}, the model's learning direction is mainly influenced by the correct pixels. However, as training advances to the \textit{`memorization phase'}, the gradient direction is gradually influenced by noisy pixels, which heavily mislead the model's learning and ultimately result in severe false negative and positive predictions. This phenomenon has also been reported in the field of classification \cite{zhang2018generalized, zhang2024cognition}. However, the manifestation of this phenomenon in COD differs from that in classification in the following aspects:  \textbf{1)} Unlike classification tasks where noise exists in only some samples, noisy pixels exist in every pseudo label in WSSCOD, and they are widespread in FSCOD training labels. \textbf{2)} There exists a spatial correlation among noisy pixels and between noisy pixels and correct pixels in the pseudo labels, and it is advantageous to use the spatial dependence to suppress noise. 

To cope with this limitation, we advocate the use of a newly proposed loss function $\mathcal{L}_{NC}$ (Noise Correction Loss) to learn to segment camouflaged objects from noisy labels.  $\mathcal{L}_{NC}$ is able to handle different learning objectives in both the early learning and the memorization phases: During the early learning phase, $\mathcal{L}_{NC}$ adapts to different fitting processes brought by different noise rates and accelerate the model's convergence to the correct pixels. Importantly, in the memorization phase, $\mathcal{L}_{NC}$ forms a unified risk gradient for different predictions, maintaining the correct learning direction on up to 50\% incorrect noisy pseudo labels, thereby aiding the model in effectively discerning visual demarcations. Furthermore, considering the prevalent noise issue in the COD training sets, $\mathcal{L}_{NC}$ also shows superior performance in WSCOD and FSCOD methods compared to their used losses. \textbf{We argue that $\mathcal{L}_{NC}$ poses a major contribution}, \textbf{as previous segmentation work has paid less attention to noisy labels, especially in the COD task}, \textbf{but where noisy labels occur more easily}. 

In summary, the main contributions of this paper are threefold:
\begin{itemize}
    \item Facing with the time-consuming and labor-intensive problem of annotating for COD tasks, we propose a cost-effective and high-performance weakly semi-supervised training scheme, and exploit the potential of box annotation as an economically accurate prompt.
    \item We propose noise correction loss to improve the model's learning of the pseudo labels generated in WSSCOD. In the early learning and memorization phases, $\mathcal{L}_{NC}$ adopts different forms to adapt to different learning objectives, ensuring correct learning of the model in the presence of noisy pixels.
    \item Compared with 16 SOTA models on four benchmark datasets, we demonstrate the superiority of the WSSCOD method, which achieves comparable performance to existing fully supervised methods with only 20\% of the annotated data for training, and proves the scalability of WSSCOD in gaining high-performance advantages with only a low-cost annotation increase.
\end{itemize}
\label{sec:intro}
\section{Weakly Semi-Supervised Camouflaged Object Detection}
\label{sec:WSSCOD}
\textbf{Task Definition.} We introduce a novel training protocol named Weakly Semi-Supervised Camouflaged Object Detection (WSSCOD), which utilizes boxes as prompts to generate high-quality pseudo labels. WSSCOD primarily leverages box annotations, complemented by a minimal amount of pixel-level annotations, to generate high-accuracy pseudo labels. Specifically, given the training set, $\mathcal{D}$, that is divided into two subsets: $\mathcal{D}_m = \{\mathcal{X}_m, \mathcal{F}_m, \mathcal{B}_m\}_{m=1}^M$, containing pixel-level annotations $\mathcal{F}_m$, box annotations $\mathcal{B}_m$ and training images $\mathcal{X}_m$. $\mathcal{D}_n = \{\mathcal{X}_n, \mathcal{B}_n\}_{n=1}^N$, containing only box annotations and images, where $M+N$ represents the number of training set. First, we train an auxiliary network, \textbf{ANet}, using the dataset $\mathcal{D}_m$, where $\mathcal{B}_m$ serves as an auxiliary prompt for camouflaged objects, and $\mathcal{F}_m$ supervises the generation of pseudo labels. Afterward, using the trained \textbf{ANet} and the dataset $\mathcal{D}_n$, we predict its pseudo labels, denoted as $\mathcal{W}_n$. Finally, we construct a weakly semi-supervised dataset $\mathcal{D}_t$ using sets $\{\mathcal{X}_m, \mathcal{F}_m\}_{m=1}^M$ and $\{\mathcal{X}_n, \mathcal{W}_n\}_{n=1}^N$, and train our proposed primary network, \textbf{PNet}, which, like other COD models, takes only images as input. The different numbers of $M$ and $N$ affect the effectiveness of \textbf{PNet}, thus, we evaluate the performance in various settings, with $M$ constituting \{1\%, 5\%, 10\%, 20\%\} and $N$ the remaining \{99\%, 95\%, 90\%, 80\%\} of the total training data $\mathcal{D}$. The resulting models are named as \textbf{PNet$_{F1}$}, \textbf{PNet$_{F5}$}, \textbf{PNet$_{F10}$}, and \textbf{PNet$_{F20}$}, respectively.

\begin{figure}[t]
            \centering
            \includegraphics[width=1\columnwidth]{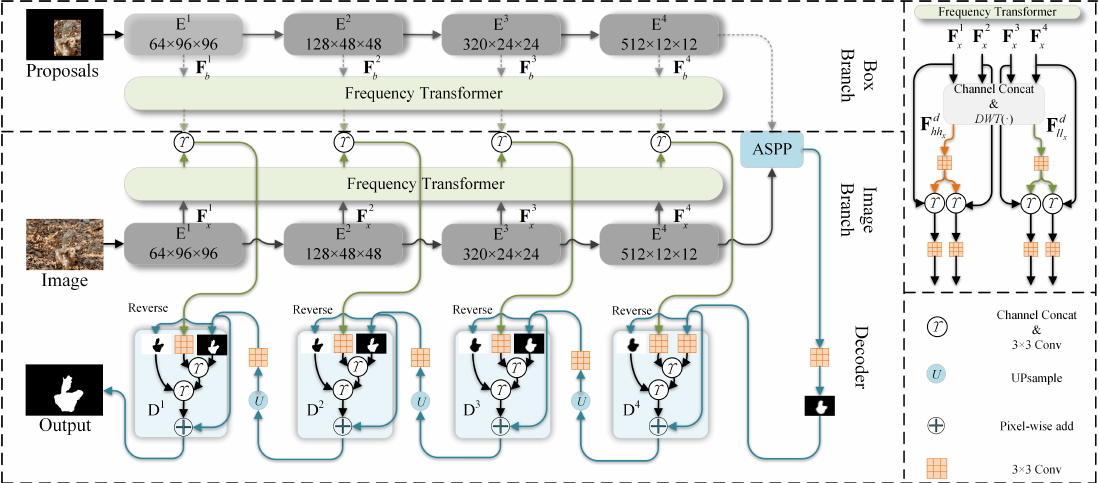}
            \caption{\textbf{Overall architecture of our proposed model for PNet and ANet}.  For \textbf{ANet}, the model consists of a box branch, an image branch, and a decoder. For \textbf{PNet}, the model consists of an image branch and a decoder.  }
		\label{fig:model}
        \end{figure}  

\subsection{Auxiliary Network}
Segmenting camouflaged objects solely based on the box often leads to inaccuracies, as the information within the box is not always reliable \cite{zhang2021deep}.  Therefore, as illustrated in Fig. \ref{fig:model}, we develop a simple and effective COD model for exploiting the complementarity between the RGB image and the proposals, but do not consider the model as a major contribution. Given an RGB image $x_m$ and an object box $b_m$, we multiply them as the input proposals $\widetilde{b}_m$ for the box branch $\textbf{BB}$ encoder, and input $x_m$ into the image branch $\textbf{IB}$ encoder.

\textbf{Encoder.} We use two ConvNeXt \cite{liu2022convnet} as encoders $E(\cdot)$ in $\textbf{ANet}$ to obtain multi-scale features for different inputs. Given the input images $\{x_m, \widetilde{b}_m\} \in \mathbb{R}^{3 \times H \times W}$, we can obtain the multi-scale features $\{\textbf{F}_x^k\}_{k=1}^4$ and $\{\textbf{F}_b^k\}_{k=1}^4$, respectively, from the branches $\textbf{IB}$ and $\textbf{BB}$ with the corresponding sizes of $\{\frac{H}{2^{k+1}}, \frac{W}{2^{k+1}}\}$ and channels $\{\textbf{C}_1, \textbf{C}_2, \textbf{C}_3, \textbf{C}_4\}$. Following established practices \cite{fang2019selective, he2023weakly, zhang2023differential, zhang2024part}, we adjust all features to the same number of channels $\textbf{C}=64$ using 3×3 convolutions for consistency across multi-level features. In addition, we apply channel concatenation to features $\textbf{F}_x^4$ and $\textbf{F}_b^4$ and then send them to ASPP \cite{chen2017rethinking} to obtain deep image representations. And the initial mask $m_{an}^4 \in \mathbb{R}^{1 \times H \times W}$ from the representations is generated through a 3×3 convolution. Subsequently, features $\textbf{F}_x^k$ and $\textbf{F}_b^k$ are respectively fed into the Frequency Transformer $\textbf{FT}$ to capture the details and deep semantics of the image, which facilitates the discriminator's recognition of camouflaged objects.

\textbf{Frequency Transformer.} Drawing inspiration from FDNet \cite{zhong2022detecting}, we employ the Discrete Wavelet Transform $DWT{(\cdot)}$ to extract both low-frequency and high-frequency components from multi-scale features. This approach is instrumental in revealing more intricate object components in camouflaged scenes by leveraging the frequency domain information $\textbf{F}^d$, enhancing the understanding of such complex visual environments. Taking $\textbf{F}_x^k$ as input
{
\small
\begin{align}
 \textbf{F}^d_{hh_x}, \textbf{F}^d_{lh_x}, \textbf{F}^d_{hl_x}, \textbf{F}^d_{ll_x} = DWT(cat(up(\textbf{F}_x^k)),
\end{align}
}where $cat{(\cdot)}$ indicates channel-wise concatenation, and $up(\cdot)$ is the up-sampling operation. In the frequency domain features, the subscripts $h$ and $l$ denote the extraction of high-frequency and low-frequency information, respectively, in the horizontal and vertical directions. To efficiently process the obtained frequency domain information and spatial domain features, we adopt an adaptive nonlinear fusion approach $\Upsilon_\omega(\textbf{F}, \textbf{F}^d)$, where $\omega$ is the learnable parameters that adjusts the degree of fusion between the two adaptively. We accomplish this fusion of shallow features $\textbf{F}_x^1$, $\textbf{F}_x^2$ with the high-frequency component $\textbf{F}^d_{hh_x}$ separately, and the integration of deep features $\textbf{F}_x^3$, $\textbf{F}_x^4$ with the low-frequency component $\textbf{F}^d_{ll_x}$. Other components such as $\textbf{F}^d_{lh_x}$ and $\textbf{F}^d_{hl_x}$ are usually not used. In $\textbf{FT}$, we represent $\Upsilon_\omega(\cdot)$ using successive convolution and channel concatenation. For easier description, the total steps in $\textbf{FT}$ are represented by $\Phi_f(\cdot)$, with $f$ specifying the type of input features. In a similar manner, $\hat{\textbf{F}}_x^k$ and $\hat{\textbf{F}}_b^k$ can be obtained through $\Phi_x(\textbf{F}_x^k)$ and $\Phi_b(\textbf{F}_b^k)$. Before proceeding with the decoding, we also perform the adaptive fusion of $\hat{\textbf{F}}_x^k$ and $\hat{\textbf{F}}_b^k$ through $\Upsilon_\omega(\cdot)$, as $\textbf{F}_c^k = \Upsilon_\omega(\hat{\textbf{F}}_x^k, \hat{\textbf{F}}_b^k)$.

\textbf{Reverse Fusion Decoder.}  We design a reverse fusion decoder to complete the convergence of multi-level features. Given the features $\{\textbf{F}_c^k\}_{k=1}^4$ and the mask $m_{an}^4$, we accomplish the fusion of multi-level features in the UNet manner. Meanwhile, we integrate a reverse mask, associating the background with difficult areas or noisy pixels in COD, amplifying the differences between them and the correct pixels, and correcting the model's learning of difficult areas, as
{
\small
\begin{equation}
p_{a n}^k=
\begin{cases}
\Upsilon_\omega\big(\Upsilon_\omega(\mathbf{F}_c^k, up(m_{a n}^4)), up(Rev(m_{a n}^4))\big) + up(m_{a n}^4), & k = 4 \\
\Upsilon_\omega\big(\Upsilon_\omega(\mathbf{F}_c^k, up(p_{a n}^{k+1})), up(Rev(p_{a n}^{k+1}))\big) + up(p_{a n}^{k+1}), & k \in\{3,2,1\},
\end{cases}
\end{equation}
}where $Rev(p_{an}^{k+1}) = -1 \times \sigma(p_{an}^{k+1}) + 1$, $\sigma$ is the sigmoid function. $\{p_{an}^k\}_{k=1}^4 \in \mathbb{R}^{1 \times H \times W}$ and $m_{an}^4$ are the predictions, in which $p_{an}^1$ is the main output of \textbf{ANet}. The decoding process is defined as $\Pi(\cdot)$, which means $p_{an}^k = \Pi(\textbf{F}_c^k, m_{an}^{4})$.

\subsection{Primary Network}
With the pretrained \textbf{ANet}, we predict the pseudo segmentation labels $\mathcal{W}_n$ by using the image set $\{\mathcal{X}_n\}_{n=1}^N$ with the corresponding box annotations $\mathcal{B}_n$, generating the training dataset $\mathcal{D}_n = \{\mathcal{X}_n, \mathcal{W}_n\}$ for the primary network $\textbf{PNet}$. Additionally, to align with existing methodologies and to maintain a consistent number of training images, we integrate the fully labeled dataset $\mathcal{D}_m = \{\mathcal{X}_m, \mathcal{F}_m\}_{m=1}^M$ into $\mathcal{D}_n$ to form the total training dataset $\{\mathcal{D}_t\}_{t=1}^{M+N}$.

In terms of model configuration, $\textbf{PNet}$ retains the same modules as $\textbf{ANet}$. However, a key difference is that $\textbf{PNet}$ employs a single-stream structure, where only the image is input. As shown in Fig. \ref{fig:model}, we use only the Image Branch and the Decoder in $\textbf{PNet}$. Specifically, in the ASPP and $\textbf{FT}$ stages, there is no channel concatenation with features from another branch. Instead, features are directly fed from the backbone network into the ASPP, and after passing through $\textbf{FT}$, they go directly into the decoder.  Given that the image $x_t$ comes from $\mathcal{D}_t$, the process of \textbf{PNet} is as $\textbf{F}_t^k = E(x_t)$ and $p_{pn}^k= \Pi(\Phi_t(\textbf{F}_t^k), ASPP(\textbf{F}_t^4))$.

\section{Noise Correction Loss}
Training $\textbf{ANet}$ with a very small amount of data poses a challenge in accurately capturing the distribution of the entire dataset, resulting in severe false negative and positive noisy pixels in the generated pseudo labels. When training on such noisy labels with traditional losses like Cross-Entropy (CE) and Intersection over Union (IoU), the bias introduced by the noisy pixels often leads to incorrect optimization directions, impacting the identification of camouflaged objects. Specifically, these losses are more sensitive to difficult pixels, which is beneficial for clean labels as it gives more bias to difficult pixels, but on noisy labels, it leads to more severe error guidance. Therefore, it is necessary to discuss the learning situation of different losses in noisy COD labels.

\subsection{Preliminaries}
We consider the learning situation of different losses on noisy labels from the perspective of gradients. Let $\{x_t, g_t\}$ be a pair of images and its noisy label in $\mathcal{D}_t$.  For any loss $\mathcal{L}$, the risk gradient of the model $\textbf{PNet}(x_t)$ can be divided as{
\small
\begin{align}
\nabla \mathcal{L}(\textbf{PNet}(x_t; \theta), g_t) = \underbrace{\nabla  \mathcal{L}(\textbf{PNet}(\widehat{x_t}), \widehat{g}_t)}_{\text{correct pixels}} +  \underbrace{\nabla  \mathcal{L}(\textbf{PNet}(\tilde{x_t}), \tilde{g}_t)}_{\text{noisy pixels}}, 
\label{equ:ce}
\end{align}
}where $\theta$ means the parameters of $\textbf{PNet}$. We consider the risk gradient by dividing the noisy label $g_t$ into two parts: correct pixels $\widehat{g}_t$ and noisy pixels $\tilde{g}_t$. When using CE or IoU loss, it is believed that the gradient values propagated by noisy pixels are greater than those from the correct pixels \cite{zhang2018generalized}. This means that the loss function introduces significant biases for noise, which are incorrect. Consequently, this leads to the model parameters $\theta$ learning in the wrong direction, ultimately affecting model's decision boundary. In contrast, as shown in Equ. \ref{equ:mae}, MAE loss does not have this issue, as it applies the same gradient to all pixels.{
\small
\begin{align}
 \frac{\partial \mathcal{L}_{MAE}}{\partial \theta} = -\nabla_\theta \textbf{PNet}g_t(x_t;\theta).
\label{equ:mae}
\end{align}
}

Moreover, MAE loss can tolerate up to 50\% noise, as the total gradient direction is still determined by the correct pixels.  However, although MAE loss is robust to noise, its constant gradient presents an optimization issue, causing it to perform poorly on challenging data, such as camouflaged images.
\begin{figure}[htbp]
            \centering
            \includegraphics[width=1\columnwidth]{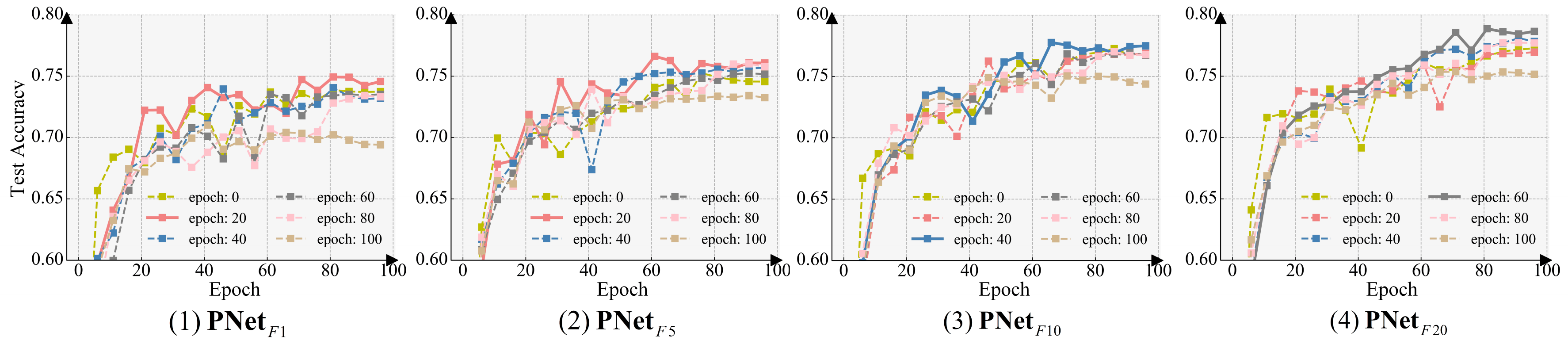}
            \caption{\textbf{Fitting analysis of different setups at different training epochs.} We train \textbf{PNet$_{F1}$}, \textbf{PNet$_{F5}$}, \textbf{PNet$_{F10}$}, and \textbf{PNet$_{F20}$} for 100 epochs and assume that the model completes the early learning phase at epochs [20,40,60,80], changing the value of $q$ from 2 to 1 to indicate the model's transition to the memorization phase.  Moreover, we set $q$ to 1 at the 0-th and 100-th epochs to use only the MAE-form and IoU-form of $\mathcal{L}_{NC}$, respectively, to validate the effectiveness of our strategy. The test metric is IoU score, and the test dataset is CAMO \cite{le2019anabranch}.} 
		\label{fig:qchange}
        \end{figure}

\subsection{$\mathcal{L}_{NC}$ Loss for Camouflaged Object Detection} 

To leverage the advantages of noise robustness offered by MAE loss and the optimization capabilities of losses such as IoU and CE losses, we propose the use of the noise correction loss $\mathcal{L}_{NC}$ in the WSSCOD task. This loss is optimized for the early learning phase and the memorization phase separately for this task, which can be calculated as follows
{
\small
\begin{align}
 \mathcal{L}_{NC} =  \frac{\sum_{i=1}^{H \times W} | p_{i} - g_{i} |^q}{ \sum_{i=1}^{H \times W} (p_{i} + g_{i}) - \sum_{i=1}^{H \times W} p_{i} \cdot g_{i}},
 \end{align}
}where $q \in [1, 2]$ is a key hyper-parameter, $p$ and $g$ are the prediction and GT. In the early learning stage, the model is required to effectively grasp the nuances of camouflaged scenes and assimilate knowledge from the correct pixels. For this purpose, we set \(q=2\), making \(\mathcal{L}_{NC}\) analogous to an IoU-form loss. In the memorization phase, the model's focus shifts towards minimizing the influence of noisy pixels and refining the decision boundary based on the gradient's guidance. At this juncture, by setting \(q=1\), \(\mathcal{L}_{NC}\) transforms into a MAE-form loss, which can guide the \textbf{PNet} to optimize in the right direction. 

Specifically, the robustness of \(\mathcal{L}_{NC}\) comes from the deterministic nature of its derivative, which does not exhibit bias towards noisy pixels. When $q=1$, the gradient of \(\mathcal{L}_{NC}\) with respect to $p_i$ is 
{
\small
\begin{align}\frac{\partial \mathcal{L}_{NC}}{\partial p_i} = \frac{\text{sign}(p_i - g_i)}{\sum_{i=1}^{H \times W} (p_i + g_i) - \sum_{i=1}^{H \times W} p_i \cdot g_i}.
\end{align}
}As we can observe, $\mathcal{L}_{NC}$ effectively combines the advantages of MAE and IoU losses: \textbf{1}) It is noise-robust as MAE, as its gradient value is the same for each predicted pixel $p_i$. \textbf{2}) Like IoU, $\mathcal{L}_{NC}$ is area-dependent, can exploit the spatial correlation between pixels, and converges faster and better than MAE.

Furthermore, an important consideration is that in different training setups, the pseudo labels predicted by $\textbf{ANet}$ are subject to varying levels of noise, leading to differences in early learning duration. Fixing the period for changing $q$ in various setups is clearly impractical, as altering it too early or too late can affect the learning of correct pixels. Consequently, we examine the effects of modifying $q$ at different epochs during different setups, as illustrated in Fig. \ref{fig:qchange}. From the figure, a key finding is that different setups require adaptation to different early learning phases; for example, changing $q$ at the 20-th epoch has a better effect for \textbf{PNet$_{F1}$}, while for \textbf{PNet$_{F20}$}, it is more effective to change at the 60-th epoch. This is because, compared to \textbf{PNet$_{F1}$}, \textbf{PNet$_{F20}$} has less noise in its training data, thus its early learning phase is longer. Moreover, using only the MAE-form or the IoU-form loss exhibits poor performance, especially in cases where noise is not corrected.  Therefore, as a result of this figure, when training \textbf{PNet$_{F1}$}, \textbf{PNet$_{F5}$}, \textbf{PNet$_{F10}$} and \textbf{PNet$_{F20}$}, we begin noise correction at the 20-th, 20-th, 40-th, and 60-th epoch, respectively.

\section{Related Work}
\subsection{Camouflaged Object Detection}
With the rapid development of deep learning technology, data-driven segmenters have achieved significant success in fully supervised COD tasks \cite{fan2020pranet, fan2020camouflaged, pang2022zoom, cong2023frequency}. PraNet \cite{fan2020pranet} introduced a parallel reverse attention mechanism, significantly improving the accuracy of detecting camouflaged objects. SINet \cite{fan2020camouflaged} mimicked the search and identification stages of animal predation to detect and locate camouflaged objects. FPNet \cite{cong2023frequency} utilized both RGB and frequency domain information for camouflaged object detection. Some weakly supervised methods \cite{zhang2020weakly, yu2021structure, he2023weakly} use points, scribbles, and point annotations to achieve low-consumption, high-precision COD. WSSA \cite{zhang2020weakly} is trained with scribble annotations and employs a gated CRF loss \cite{obukhov2019gated} to enhance object detection accuracy. SCOD \cite{he2023weakly} introduced a novel consistency loss to ensure the agreement of individual prediction maps, leveraging the loss from an internal viewpoint. However, weakly supervised methods still remain a significant challenge in COD tasks, as the high similarity of camouflaged images prevents these methods from distinguishing between foreground and background. Therefore, unlike previous fully-supervised or weakly-supervised methods, we propose a new learning strategy, WSSCOD, which aims to achieve high-performance COD with an economical and labor-saving labeling approach.

\subsection{Learning with Noisy Label}
Deep learning algorithms' remarkable performance heavily relies on large-scale, high-quality human annotations, obtaining which is extremely costly and time-consuming. Cheaper annotation methods like web scraping and weakly supervised methods offer an economical and efficient way to gather labels, but the noise in these labels is inevitable. Learning with noisy labels aims to provide various strategies to tackle this challenging issue, such as robust loss design, noise transition matrices, and sample selection. Zhang et al. \cite{zhang2018generalized} introduced a generalized cross-entropy loss, which allows training with noisy labels by down weighting the contribution of noisy samples. Patrini et al. \cite{patrini2017making} proposed a method to estimate the noise transition matrix, which represents the probabilities of true labels being flipped to other labels, improving model training under label noise. Han et al. \cite{han2018co} developed a co-teaching approach where two networks teach each other what they have learned, effectively reducing the impact of noisy labels by selecting clean samples during training. Noise is unavoidable in the WSSCOD task, and we construct $\mathcal{L}_{NC}$ to facilitate the model's learning of correct pixels as well as the correction of noisy pixels.

\section{Experiments}
\label{sec:exp}

\subsection{Experimental Settings}
\textbf{Datasets and Metrics.} In COD task, four primary datasets serve as benchmarks: CAMO \cite{le2019anabranch}, COD10K \cite{fan2020camouflaged}, CHAMELEON \cite{skurowski2018animal}, and NC4K \cite{lv2021simultaneously}, containing 250, 2026, 4121, and 76 image pairs, respectively. The training set consists of 4040 pairs, with 1000 from CAMO and 3040 from COD10K. Within the WSSCOD framework, we leverage box annotations and a subset of fully annotated images. The approach includes four setups, partitioning the training set randomly into subsets of \{1\%, 5\%, 10\%, 20\%\} of images with full and box annotations, and the remaining \{99\%, 95\%, 90\%, 80\%\} with only box annotations.

Following the methodologies established in \cite{pang2022zoom, lv2021simultaneously}, four essential metrics are adopted for an in-depth evaluation of model performance: mean absolute error ($\mathcal{M}$), E-measure ($E_{\phi}$) \cite{fan2021cognitive}, F-measure ($F_{\beta}$) \cite{margolin2014evaluate}, and S-measure ($S_{\alpha}$) \cite{fan2017structure}. 

\textbf{Implementation Details.} \textit{\textbf{For models}}: Following the selection practices of existing COD methods, we choose the SOTA backbone network PVTv2-B4 \cite{wang2022pvt} as the encoder for PNet to demonstrate the effectiveness of our method. 
However, for ANet, due to the Transformer model's weak performance on small-scale data \cite{dosovitskiy2020image}, we opt for ConvNeXt-B \cite{liu2022convnet} as ANet's encoder. The weights of these backbone networks are pretrained on ImageNet \cite{deng2009imagenet}.
\textit{\textbf{For data}}: To enhance the model's robustness, we apply data augmentation techniques such as random cropping, random blurring, random brightness adjustments, and random flipping to the training images. Subsequently, the images are resized to 384×384 before being fed into the WSSCOD framework. 
\textit{\textbf{For training}}: We use Adam optimizer \cite{kingma2014adam} to update model parameters, with both ANet and PNet trained for 100 epochs. The initial learning rate is set to 1e-7, linearly warmed up over 10 epochs to 1e-4, followed by cosine annealing down to 1e-7 while training. All random factors, including data selection and the training process, are fixed using seed 2024 to ensure the reproducibility of the model. Besides the NC loss, we also use DICE loss, similar to FEDER \cite{he2023camouflaged}  and BGNet \cite{sun2022boundary}, to assist the model in learning the object boundaries.

\textbf{Comparison Methods.} 
To validate the effectiveness of the proposed WSSCOD method, we construct a comparison of it with several recent SOTA methods, including the WSCOD methods WSSA \cite{zhang2020weakly}, SCWS \cite{yu2021structure}, TEL \cite{liang2022tree}, SCOD \cite{he2023weakly}, the fully supervised methods SINet \cite{fan2020camouflaged}, FEDER \cite{he2023camouflaged}, SINetv2 \cite{fan2021concealed}, BSA-Net \cite{zhu2022can}, BGNet \cite{sun2022boundary}, CamoFormer \cite{yin2022camoformer}, FSPNet \cite{huang2023feature}, HitNet \cite{hu2023high}, MSCAF-Net \cite{liu2023mscaf}, and the prompt based methods SAM \cite{kirillov2023segment}, SAM with point prompt (SAM-P), SAM with box prompt (SAM-B). The results of these methods come from public data or are generated by models retrained with the released code. The comparison results are shown in Table \ref{tab:comparison}, and the qualitative comparison is in Fig \ref{fig:compare}.

\subsection{Performance Comparision with SOTAs}
\begin{table}[t]
\renewcommand{\arraystretch}{0.9}
\Large
  \centering
  \caption{\textbf{Performance comparison with SOTA methods}.  $\downarrow$ ($\uparrow $) means that the lower (higher) is better. Our methods are marked with \cb, where \textbf{PNet}$_{F20}^\dag$ indicates the performance achieved using an additional training set of 6665 images with only box annotations from ACOD \cite{song2023camouflaged} and CoCOD \cite{zhang2023collaborative} datasets. $\mathcal{S}$, $\mathcal{B}$, $\mathcal{P}$ and $\mathcal{F}$ represent the scribble, box, point, and pixel-level annotations separately.}
    \resizebox{\textwidth}{!}{\setlength{\tabcolsep}{0.5mm}{
    \begin{tabular}{l|c|c|cccc|cccc|cccc|cccc}
    \toprule
     \rowcolor{tabtitle}
    & & & \multicolumn{4}{c|}{\textbf{CAMO} (250)} & \multicolumn{4}{c|}{\textbf{CHAMELEON} (76)} & \multicolumn{4}{c|}{\textbf{COD10K} (2026)} & \multicolumn{4}{c}{\textbf{NC4K} (4121)} \\
\rowcolor{tabtitle}
\multirow{-2}{*}{\textbf{Model}} & \multirow{-2}{*}{\textbf{Encoder}} & \multirow{-2}{*}{\textbf{Annotation}}  & $\mathcal{M}\downarrow$ & $E_{\phi}\uparrow$ & $F_{\beta}\uparrow$ & $S_{\alpha}\uparrow$ & $\mathcal{M}\downarrow$ & $E_{\phi}\uparrow$ & $F_{\beta}\uparrow$ & $S_{\alpha}\uparrow$ & $\mathcal{M}\downarrow$ & $E_{\phi}\uparrow$ & $F_{\beta}\uparrow$ & $S_{\alpha}\uparrow$ & $\mathcal{M}\downarrow$ & $E_{\phi}\uparrow$ & $F_{\beta}\uparrow$ & $S_{\alpha}\uparrow$ \\
    \midrule
    \multicolumn{18}{c}{\textit{\textbf{Weakly Supervised Methods}}} \\
    \midrule
   WSSA$^{20}$ \cite{zhang2020weakly}  & ResNet-50 &  $\mathcal{S}$ 100\% & 0.118  & 0.686  & 0.632  & 0.696  & 0.067  & 0.860  & 0.719  & 0.782  & 0.071  & 0.770  & 0.598  & 0.684  &  0.091  & 0.779  & 0.528  & 0.761  \\
    SCWS$^{21}$ \cite{yu2021structure} & ResNet-50& $\mathcal{S}$ 100\% & 0.102 & 0.658 & 0.651 & 0.713  & 0.053 & 0.881 & 0.721 & 0.792  & 0.055 & 0.805 & 0.644 & 0.710  & 0.073 & 0.814 & 0.713 & 0.784  \\
    TEL$^{22}$ \cite{liang2022tree} & ResNet-50& $\mathcal{B}$ 100\% & 0.104 & 0.681 & 0.654 & 0.717 & 0.073 & 0.827 & 0.706 & 0.785 & 0.057 & 0.801 & 0.659 & 0.724 & 0.075 & 0.799 & 0.701 & 0.782  \\
    SCOD$^{23}$ \cite{he2023weakly} & ResNet-50& $\mathcal{P}$ 100\% & 0.129 & 0.688 & 0.592 & 0.663  & 0.092 & 0.746 & 0.692 & 0.725  & 0.060 & 0.802 & 0.628 & 0.711  & 0.080 & 0.796 & 0.686 & 0.758  \\
\midrule
\multicolumn{18}{c}{\textit{\textbf{Fully Supervised Methods}}} \\
    \midrule
   SINet$^{20}$ \cite{fan2020camouflaged}    & ResNet-50           & $\mathcal{F}$ 100\%                  &  0.100  &  0.771  &  0.675  &  0.751 &  0.043  &  0.891  &  0.787  &  0.869 &  0.051  &  0.806  &  0.634  &  0.771 &  0.058  &  0.871  &  0.769  &  0.808 \\
  FEDER$^{23}$ \cite{he2023camouflaged}   & ResNet-50       & $\mathcal{F}$ 100\%                  & 0.071 & 0.898 & 0.781 & 0.802 & 0.030 & 0.959 & 0.851 & 0.887 & 0.032 & 0.905 & 0.751 & 0.822 & 0.044 & 0.915 & 0.824 & 0.847 \\
    SINetv2$^{21}$   \cite{fan2021concealed}   & Res2Net-50       & $\mathcal{F}$ 100\%              & 0.070 & 0.895 & 0.782 & 0.820 & 0.030 & 0.961 & 0.835 & 0.888 & 0.037 & 0.906 & 0.718 & 0.815 & 0.048 & 0.914 & 0.805 & 0.847 \\
BSA-Net$^{22}$  \cite{zhu2022can}    & Res2Net-50       & $\mathcal{F}$ 100\%              &  0.079  &  0.851  &  0.763  &  0.794 &  0.026  &  0.946  &  0.856  &  0.896  &  0.034  &  0.891  &  0.738  &  0.818  &  0.048  &  0.897  &  0.808  &  0.841 \\
BGNet$^{22}$  \cite{sun2022boundary}    & Res2Net-50       & $\mathcal{F}$ 100\%               &  0.073  &  0.870  &  0.789  &  0.812 &  0.027  &  0.943  &  0.857  &  0.901 &  0.033  &  0.901  &  0.753  &  0.831 &  0.044  &  0.907  &  0.820  &  0.851 \\
CamoFormer$^{22}$ \cite{yin2022camoformer}   & PVTv2-B4       & $\mathcal{F}$ 100\%              &0.046  & 0.929  & 0.854  & 0.872 &  0.022  &  0.957  &  0.880  &0.909  &  0.023  &  0.932  &  0.811  &  0.869 &  0.030  &  0.939  &  0.868  &  0.891 \\
FSPNet$^{23}$  \cite{huang2023feature}    & ViT-B16       & $\mathcal{F}$ 100\%              &  0.050  &  0.899  &  0.830  &  0.856 &  0.022  &  0.942  &  0.865  &  0.908 &  0.026  &  0.895  &  0.769  &  0.851 &  0.035  &  0.915  &  0.843  &  0.879 \\
HitNet$^{23}$   \cite{hu2023high}    & PVTv2-B2       & $\mathcal{F}$ 100\%              &  0.055  &  0.906  &  0.831  &  0.849 &  0.019  &  0.966  &  0.898  &  0.921 &  0.023  &  0.935  &  0.823  &  0.871 &  0.037  &  0.926  &  0.853  &  0.875 \\
MSCAF-Net$^{23}$  \cite{liu2023mscaf}    & PVTv2-B2       & $\mathcal{F}$ 100\%              &  0.046  &  0.929  &  0.852  &  0.873 &  0.022  &  0.958  &  0.874  &  0.911 &  0.024  &  0.927  &  0.798  &  0.865 &  0.032  &  0.934  &  0.860  &  0.887 \\
\midrule
\multicolumn{18}{c}{\textit{\textbf{Prompt Based Methods}}} \\
    \midrule
SAM$^{23}$ \cite{kirillov2023segment}    & ViT-H           & -                  &  0.209  &  0.304  &  0.039  &  0.394 &  0.157  &  0.276  &  0.017  &  0.418 &  0.111  &  0.315  &  0.018  &  0.445 &  0.186  &  0.338  &  0.040  &  0.406 \\
  SAM-P$^{23}$   \cite{kirillov2023segment}   & ViT-H           & -                 &  0.126  &  0.653  &  0.595  &  0.658 &  0.068  &  0.737  &  0.666  &  0.731 &  0.084  &  0.725  &  0.613  &  0.706 &  0.108  &  0.711  &  0.637  &  0.698 \\
  SAM-B$^{23}$ \cite{kirillov2023segment}   & ViT-H         & -                 &  0.139  &  0.495  &  0.346  &  0.535 &  0.121  &  0.467  &  0.276  &  0.524 &  0.073  &  0.433  &  0.218  &  0.534 &  0.110  &  0.510  &  0.348  &  0.558 \\
  \midrule
\multicolumn{18}{c}{\textit{\textbf{Weakly Semi-Supervised Methods}}} \\
    \midrule
\gr
\cb \textbf{PNet$_{F1}$} & PVTv2-B4  & $\mathcal{F}$ 1\% $+$ $\mathcal{B}$ 99\% &  0.051  &  0.922  &  0.835  &  0.852  &  0.038  &  0.921  &  0.812  &  0.847 &  0.031  &  0.903  &  0.745  &  0.828 &  0.037  &  0.926  &  0.831  &  0.864 \\
\gr
\cb \textbf{PNet$_{F5}$} & PVTv2-B4  & $\mathcal{F}$ 5\% $+$ $\mathcal{B}$ 95\% &  0.050  &  0.924  &  0.845  &  0.857 &  0.032  &  0.943  &  0.821  &  0.865 &  0.027  &  0.921  &  0.771  &  0.845&  0.034  &  0.934  &  0.844  &  0.874 \\
\gr
\cb \textbf{PNet$_{F10}$} & PVTv2-B4  & $\mathcal{F}$ 10\% $+$ $\mathcal{B}$ 90\% &  0.048  &  0.925  &  0.841  &  0.861 &  0.028  &  0.949  &  0.830  &  0.878 &  0.024  &  0.927  &  0.782  &  0.855 &  0.032  &  0.937  &  0.848  &  0.880 \\
\gr
\cb \textbf{PNet$_{F20}$} & PVTv2-B4  & $\mathcal{F}$ 20\% $+$ $\mathcal{B}$ 80\% &  \pmb{0.043}  &  \pmb{0.934}  &  \pmb{0.856}  &  \pmb{0.872} &  \pmb{0.024}  &  \pmb{0.954}  &  \pmb{0.861}  &  \pmb{0.892}  &  \pmb{0.023}  &  \textbf{0.932}  &  \pmb{0.792}  &  \pmb{0.860} &  \pmb{0.031}  &  \pmb{0.940}  &  \pmb{0.857}  &  \pmb{0.885} \\
\midrule
\gr
\cb \textbf{PNet$_{F20}^\dag$} & PVTv2-B4  & $\mathcal{F}$ 20\% $+$ $\mathcal{B}$ 240\% &    \pmb{\textit{0.039}} & \pmb{\textit{0.942}} & \pmb{\textit{0.870}} & \pmb{\textit{0.882}} & \pmb{\textit{0.021}} & \pmb{\textit{0.964}} & \pmb{\textit{0.886}} & \pmb{\textit{0.908}} & \pmb{\textit{0.016}} & \pmb{\textit{0.960}} & \pmb{\textit{0.857}} & \pmb{\textit{0.901}} & \pmb{\textit{0.024}} & \pmb{\textit{0.958}} & \pmb{\textit{0.888}} & \pmb{\textit{0.906}} \\
    \bottomrule
    \end{tabular}}}%
  \label{tab:comparison}%
\end{table}%

\textbf{Quantitative Evaluation.} Table \ref{tab:comparison} provides a comprehensive quantitative comparison between our proposed WSSCOD and 16 other COD models, using various training strategies such as WSCOD methods, FSCOD methods and prompt based methods. As shown in the table, our \textbf{PNet$_{F1}$} (with pixel-level annotations for only 40 images) outperforms weakly supervised methods, achieving an average improvement of 79.6\%, 16.3\%, 18.1\%, and 16.0\% in $\mathcal{M}$, $E_{\phi}$, $F_{\beta}$, and $S_{\alpha}$ metrics, respectively, compared to SCWS \cite{yu2021structure}. Compared to the fully supervised SOTA method CamoFormer \cite{yin2022camoformer}, our \textbf{PNet$_{F20}$} exhibits comparable performance on the four datasets, with a gap of less than 1\%, while requiring only about 1/5 of their annotation effort (pixel-level annotations for just 800 images). Compared to SAM \cite{kirillov2023segment}, our method demonstrates significant advantages, even against the box or point-prompted SAM. More importantly, our WSSCOD method is scalable, and incremental training only requires box annotations, as demonstrated by \textbf{PNet$_{F20}^\dag$}. By training with an additional 6665 images with only box annotations, we achieve higher performance improvements. Compared to \textbf{PNet$_{F20}$}, \textbf{PNet$_{F20}^\dag$} shows an apparent improvement on the COD10K and NC4K datasets, with improvements of 35\%, 2.4\%, 5.8\%, and 3.6\% in the four metrics. Overall, the results of \textbf{PNet} show the success of the WSSCOD method in breaking the time-consuming labeling process, providing new insights for COD task.

\begin{wrapfigure}{l}{0.5\textwidth}
\centering
\includegraphics[width=0.99\linewidth]{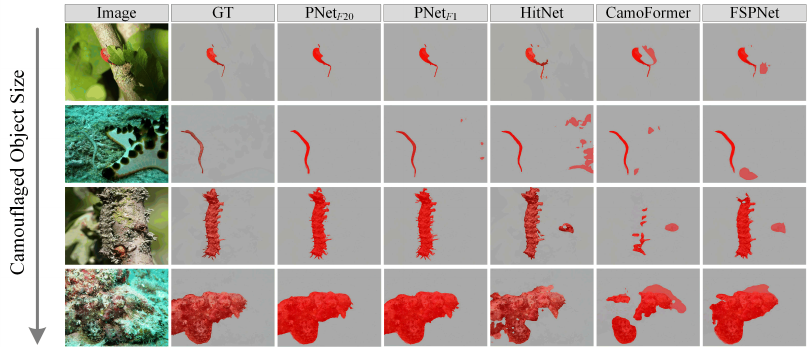}
\caption{\textbf{Visualization comparison of ours and SOTA methods}. Please zoom in to view.}
\label{fig:compare}
\end{wrapfigure}

\textbf{Qualitative Evaluation.} We select some representative COD scenes for visual comparison in Fig. \ref{fig:compare}. These scenes reflect various scenarios, including various types of camouflaged objects of different sizes and dimensions. From these results, our methods excel in preserving semantic accuracy and ensuring the integrity of fine edges, surpassing other models that may suffer from over-prediction, ambiguous details, and missing edges. It is worth noting that our results are learned from more noisy labels.

\subsection{Ablation Experiments}
The following ablation study validates the innovations of this research, particularly the WSSCOD strategy and the $\mathcal{L}_{NC}$ loss function. 

\begin{figure}[t]
\centering
\begin{minipage}{0.46\textwidth}
\centering
\includegraphics[width=5cm]{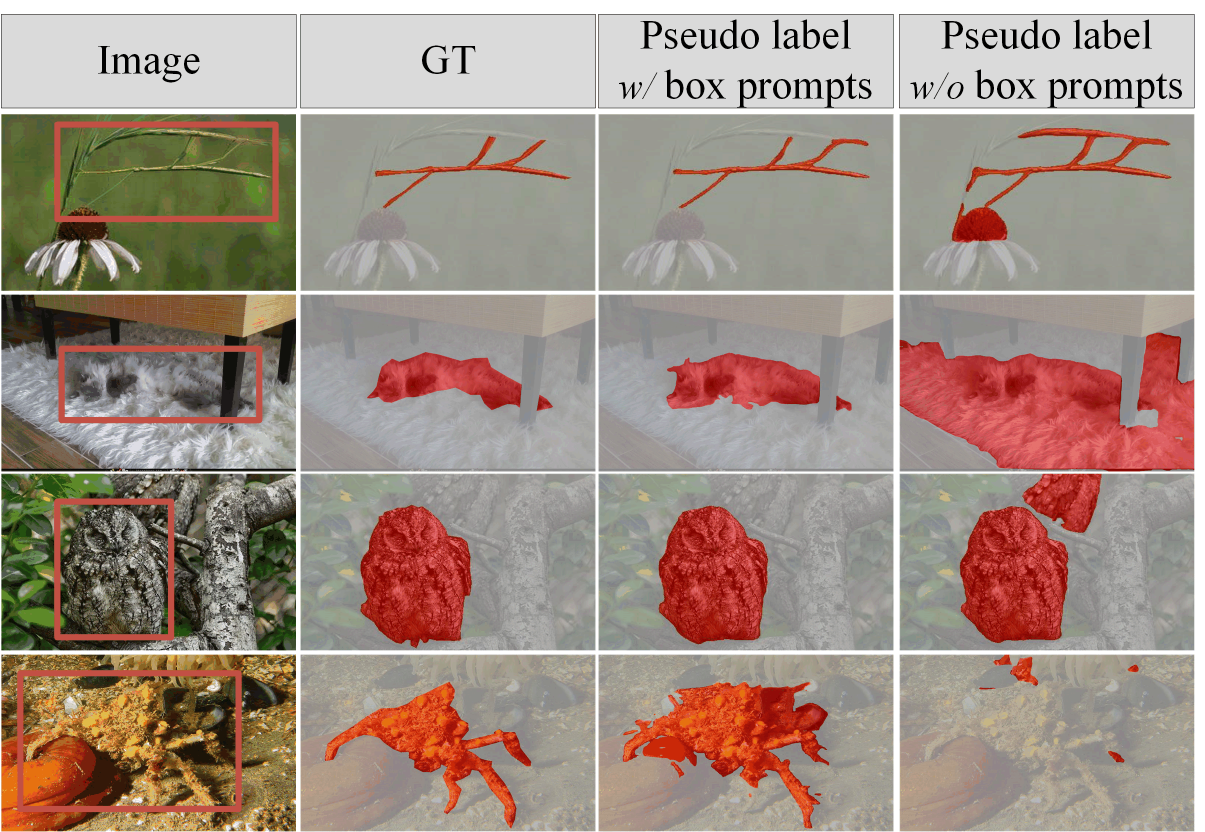}
\caption{\textbf{Pseudo label examples obtained by training \textbf{ANet} with and without boxes}. The third and fourth columns show pseudo labels generated with and without the box as prompts. We represent the box prompts used on the images.}
\label{fig:box}
\end{minipage}
\begin{minipage}{0.52\textwidth}
\centering
\includegraphics[width=6cm]{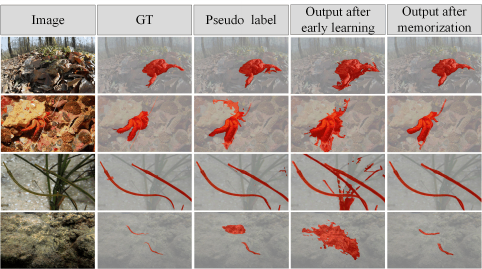}
\caption{\textbf{Visual examples of the effects of supervising in early learning and memorization phases}. The fourth column represents the model's output during the early learning phase, and the fifth column shows the model's output during the memorization stage with noise correction.}
\label{fig:effwsscod}
\end{minipage}
\end{figure}

\textbf{Effect of Box Prompts.} We select the box annotations as additional prompts and provide a performance comparison with other annotation types, such as points and scribbles, in Table \ref{tab:box}. For point annotations, denoted as $\mathcal{P}$, we use the method recommended by \cite{kim2023devil}, namely MaskRefineNet, to refine the output. The processing method for scribble annotations, denoted as $\mathcal{S}$, is consistent with our approach for box annotations, and the test dataset used is $\mathcal{D}_n$. According to this table, the improvement with box annotations as prompts is significant, surpassing both scribble and point annotations by more than 7.2\% (0.802 vs. 0.860 in $F_{\beta}$) and achieving a 14.5\% improvement in performance (0.751 vs. 0.860 in $F_{\beta}$) compared to methods without any prompts. Fig. \ref{fig:box} illustrates how box prompts refine the quality of pseudo labels by preventing model misjudgments and enhancing the distinction between object and background. In total, boxes are effective in COD tasks, as they greatly slow down the pressure on the model to detect in camouflaged scenes.

\textbf{Effect of WSSCOD}. We introduce WSSCOD as an innovative training strategy and provide a comparative analysis with other methods, such as semi-supervised COD and FSCOD, in Table \ref{tab:wsscod}. This table includes comparisons of training with 20\% (line \textcircled{1}) and 100\% (line \textcircled{2}) of fully supervised data, as well as training with a combination of 20\% fully supervised data and 80\% unlabeled data $\mathcal{U}$ (line \textcircled{3}), juxtaposed against our method (line \textcircled{4}). Observations from \textcircled{1}, \textcircled{2}, and \textcircled{4} indicate that leveraging additional data can significantly boost model performance. Furthermore, by incorporating boxes as masks, our method concentrates more effectively on the segmentation of camouflaged objects, achieving results comparable to those obtained with 100\% full pixel-level data, yet at a considerably lower annotation cost than \textcircled{2}. Compared to \textcircled{3}, while the semi-supervised approach is more time- and labor-efficient, it is clearly far from our WSSCOD method and is difficult to scale more efficiently.
\begin{table}
\parbox{.45\linewidth}{
\centering
\caption{\textbf{Effect of box prompts}. We use different types of annotations to assist $\textbf{ANet}$ training. The test dataset is the rest 80\% of the training data $\mathcal{D}_n$.}
\scalebox{0.80}{\setlength{\tabcolsep}{1mm}{\begin{tabular}{l|cccc}
    \toprule
    Label Types & $\mathcal{M}\downarrow$ & $E_{\phi}\uparrow$ & $F_{\beta}\uparrow$ & $S_{\alpha}\uparrow$ \\
    \midrule
    $\mathcal{F}$ 20\%    &  0.037  &  0.894  &  0.751  &  0.824 \\
    $\mathcal{F}$ 20\% + $\mathcal{P}$ 20\% & 0.035 & 0.900 & 0.755 & 0.834 \\
    $\mathcal{F}$ 20\% + $\mathcal{S}$ 20\% & 0.030 & 0.921 & 0.802 & 0.857 \\
    \midrule
    \gr
     $\mathcal{F}$ 20\% + $\mathcal{B}$ 20\% &  \textbf{0.021}  &  \textbf{0.960}  &  \textbf{0.860}  &  \textbf{0.900} \\
    \bottomrule
    \end{tabular}}}%
\label{tab:box}
}
\hfill
\parbox{.450\linewidth}{
\centering
\caption{\textbf{Effect of the WSSCOD}. We use different types of annotations to train $\textbf{PNet}$. The test dataset is COD10K.}
\scalebox{0.80}{\begin{tabular}{l|cccc}
    \toprule
    Label Types & $\mathcal{M}\downarrow$ & $E_{\phi}\uparrow$ & $F_{\beta}\uparrow$ & $S_{\alpha}\uparrow$ \\
    \midrule
    \textcircled{1} $\mathcal{F}$ 20\% &  0.031  &  0.892  &  0.764  &  0.847 \\
    \textcircled{2} $\mathcal{F}$ 100\% & \textbf{0.022} & 0.922 & \textbf{0.808} & \textbf{0.877} \\
    \textcircled{3} $\mathcal{F}$ 20\% + $\mathcal{U}$ 80\% & 0.029 & 0.898 & 0.771 & 0.850 \\
    \midrule
   \gr
   \textcircled{4} $\mathcal{F}$ 20\% + $\mathcal{B}$ 80\% &  0.023  &  \textbf{0.932}  &  0.792  &  0.860 \\
    \bottomrule
    \end{tabular}}%
\label{tab:wsscod}
}
\end{table}

\textbf{Effect of Noise Correction Loss.} One key innovation in our work is the noise correction loss ($\mathcal{L}_{NC}$), which is designed to enhance the model's robustness to noisy labels. To evaluate its effectiveness, we compare the performance of models trained with various loss functions in Table \ref{tab:losses}. Consistent with the conclusions in Fig. \ref{fig:qchange}, the sensitivity of CE and IoU/IoU-form $\mathcal{L}_{NC}^{q=2.0}$ losses to noise leads to their poor performance (0.780/0.778 vs. 0.792 in $F_{\beta}$). Using only the MAE-form of loss also does not yield optimal performance (0.778 vs. 0.792 in $F_{\beta}$). GCE is a well-known work in noise learning, but its performance in the COD task is not satisfactory (4\% gap in $F_{\beta}$), highlighting the differences between the two tasks. Fig. \ref{fig:effwsscod} illustrates the \textbf{PNet}'s outputs, demonstrating how \(\mathcal{L}_{NC}\) corrects noise in the memorization phase, enabling precise capture of object's details. This balanced approach ensures optimal learning from noisy labels, showcasing the $\mathcal{L}_{NC}$'s ability to enhance model accuracy and reliability.
\begin{table}
\parbox{.45\linewidth}{
\centering
\caption{\textbf{Effect of noise correction loss}. $\mathcal{L}_{NC}^{q=2.0}$ or $\mathcal{L}_{NC}^{q=1.0}$ means $q$ remains constant throughout training. }
\scalebox{0.80}{\setlength{\tabcolsep}{1mm}{\begin{tabular}{c|cccc}
    \toprule
    Losses & $\mathcal{M}\downarrow$ & $E_{\phi}\uparrow$ & $F_{\beta}\uparrow$ & $S_{\alpha}\uparrow$ \\
    \midrule
    CE + IoU   &  0.027  &  0.910  &  0.780  &  0.844 \\
    CE + $\mathcal{L}_{NC}^{q=2.0}$ &  0.027  &  0.912  &  0.778  &  0.849 \\
    $\mathcal{L}_{NC}^{q=1.0}$ &  0.024  &  0.922  &  0.780  &  0.855 \\
    GCE \cite{zhang2018generalized} &  0.029  &  0.900  &  0.759  &  0.835 \\
    \midrule
    \gr
     $\mathcal{L}_{NC}$  &  \textbf{0.023}  &  \textbf{0.932}  &  \textbf{0.792}  &  \textbf{0.860} \\
    \bottomrule
    \end{tabular}}}%
\label{tab:losses}
}
\hfill
\parbox{.450\linewidth}{
\centering
\caption{\textbf{Universality of noise correction loss}. We retrain the fully/weakly supervised COD methods with/without our loss.}
\scalebox{0.80}{\begin{tabular}{c|c|cccc}
    \toprule
    Model & Losses & $\mathcal{M}\downarrow$ & $E_{\phi}\uparrow$ & $F_{\beta}\uparrow$ & $S_{\alpha}\uparrow$ \\
    \midrule
    \multirow{2}[2]{*}{SINetv2} & w/o $\mathcal{L}_{NC}$ & \multicolumn{1}{r}{0.037} & \multicolumn{1}{r}{0.887} & \multicolumn{1}{r}{0.719} & \multicolumn{1}{r}{0.814} \\
          & \cellcolor[rgb]{ .906,  .902,  .902} w/ $\mathcal{L}_{NC}$ &  \cellcolor[rgb]{ .906,  .902,  .902} \textbf{0.033}  &  \cellcolor[rgb]{ .906,  .902,  .902} \textbf{0.896}  & \cellcolor[rgb]{ .906,  .902,  .902} \textbf{0.734}  & \cellcolor[rgb]{.906,  .902,  .902} \textbf{0.814} \\
    \midrule
    \multirow{2}[2]{*}{SCOD} & w/o $\mathcal{L}_{NC}$ & 0.060 & 0.802 & 0.628 & 0.711 \\
          & \cellcolor[rgb]{.906,  .902,  .902} w/ $\mathcal{L}_{NC}$ & \cellcolor[rgb]{.906,  .902,  .902}\textbf{0.045} & \cellcolor[rgb]{.906,  .902,  .902}\textbf{0.853} & \cellcolor[rgb]{.906,  .902,  .902}\textbf{0.665} & \cellcolor[rgb]{.906,  .902,  .902}\textbf{0.759} \\
    \bottomrule
    \end{tabular}%
  \label{tab:addlabel}}%
	\label{tab:univer}
}
\end{table}

\textbf{Universality of Noise Correction Loss.} We posit that \(\mathcal{L}_{NC}\) is versatile, as noise is commonly present in both WSCOD and FSCOD training sets. Thus, we conduct experiments on the fully supervised model SINetv2 and the weakly supervised model SCOD, modifying their training loss functions to our \(\mathcal{L}_{NC}\) during training, and changing the parameter \(q\) to 1 in the later stages of training. The results are shown in Table \ref{tab:univer}. It can be seen that both models have achieved effective performance improvements, where the FSCOD method SINetv2 is improved by 12.1\%, 1\%, and 2.1\% in $\mathcal{M}$, $E_{\phi}$ and $F_{\beta}$ metrics except for $S_{\alpha}$. The SCOD is also improved by 33.3\%, 7.0\%, 5.9\%, and 6.8\% in the four metrics.

\section{Conclusion and Discussion}
\textbf{Conclusion.} We proposed WSSCOD to achieve low-cost, high-performance COD. Moreover, to address the issue of noisy pseudo labels generated by \textbf{ANet}, we introduced \(\mathcal{L}_{NC}\) to achieve gradient consistency under noisy pixels. Our method requires only 20\% of full annotations to reach the SOTA performance. 

\textbf{Limitation.} One major limitation of the proposed method is that the accuracy of box annotation has a bit of impact on the final results, which is similar to the issue of multimodal bias. Specifically, in \textbf{ANet}, we use channel concatenation to fuse the dual branches instead of employing overly complex fusion strategies. As we aim to keep it simple, treating it as a baseline model. Actually, a better fusion strategy could mitigate the impact of incorrect boxes and improve performance. Another limitation is that WSSCOD is two-stage, which is cumbersome, and a direction for subsequent research.

\textbf{Acknowledgments.} This work was funded by  STI 2030—Major Projects under grant 2022\ ZD0209600, National Natural Science Foundation of China  62201058 and 6180612.

%
%
\bibliographystyle{splncs04}
\bibliography{main}

\begin{thebibliography}{10}
\providecommand{\url}[1]{\texttt{#1}}
\providecommand{\urlprefix}{URL }
\providecommand{\doi}[1]{https://doi.org/#1}

\bibitem{chen2017rethinking}
Chen, L.C., Papandreou, G., Schroff, F., Adam, H.: Rethinking atrous
  convolution for semantic image segmentation. arXiv preprint arXiv:1706.05587
  (2017)

\bibitem{chibane2022box2mask}
Chibane, J., Engelmann, F., Anh~Tran, T., Pons-Moll, G.: Box2mask: Weakly
  supervised 3d semantic instance segmentation using bounding boxes. In:
  European Conference on Computer Vision. pp. 681--699. Springer (2022)

\bibitem{cong2023frequency}
Cong, R., Sun, M., Zhang, S., Zhou, X., Zhang, W., Zhao, Y.: Frequency
  perception network for camouflaged object detection. In: Proceedings of the
  31st ACM International Conference on Multimedia. pp. 1179--1189 (2023)

\bibitem{deng2009imagenet}
Deng, J., Dong, W., Socher, R., Li, L.J., Li, K., Fei-Fei, L.: Imagenet: A
  large-scale hierarchical image database. In: 2009 IEEE conference on computer
  vision and pattern recognition. pp. 248--255. Ieee (2009)

\bibitem{dosovitskiy2020image}
Dosovitskiy, A., Beyer, L., Kolesnikov, A., Weissenborn, D., Zhai, X.,
  Unterthiner, T., Dehghani, M., Minderer, M., Heigold, G., Gelly, S., et~al.:
  An image is worth 16x16 words: Transformers for image recognition at scale.
  arXiv preprint arXiv:2010.11929  (2020)

\bibitem{fan2017structure}
Fan, D.P., Cheng, M.M., Liu, Y., Li, T., Borji, A.: Structure-measure: A new
  way to evaluate foreground maps. In: Proceedings of the IEEE international
  conference on computer vision. pp. 4548--4557 (2017)

\bibitem{fan2021concealed}
Fan, D.P., Ji, G.P., Cheng, M.M., Shao, L.: Concealed object detection. IEEE
  transactions on pattern analysis and machine intelligence  \textbf{44}(10),
  6024--6042 (2021)

\bibitem{fan2021cognitive}
Fan, D.P., Ji, G.P., Qin, X., Cheng, M.M.: Cognitive vision inspired object
  segmentation metric and loss function. Scientia Sinica Informationis
  \textbf{6}(6) (2021)

\bibitem{fan2020camouflaged}
Fan, D.P., Ji, G.P., Sun, G., Cheng, M.M., Shen, J., Shao, L.: Camouflaged
  object detection. In: Proceedings of the IEEE/CVF conference on computer
  vision and pattern recognition. pp. 2777--2787 (2020)

\bibitem{fan2020pranet}
Fan, D.P., Ji, G.P., Zhou, T., Chen, G., Fu, H., Shen, J., Shao, L.: Pranet:
  Parallel reverse attention network for polyp segmentation. In: International
  conference on medical image computing and computer-assisted intervention. pp.
  263--273. Springer (2020)

\bibitem{fang2019selective}
Fang, Y., Chen, C., Yuan, Y., Tong, K.y.: Selective feature aggregation network
  with area-boundary constraints for polyp segmentation. In: Medical Image
  Computing and Computer Assisted Intervention--MICCAI 2019: 22nd International
  Conference, Shenzhen, China, October 13--17, 2019, Proceedings, Part I 22.
  pp. 302--310. Springer (2019)

\bibitem{han2018co}
Han, B., Yao, Q., Yu, X., Niu, G., Xu, M., Hu, W., Tsang, I.W., Sugiyama, M.:
  Co-teaching: Robust training of deep neural networks with extremely noisy
  labels. In: Advances in Neural Information Processing Systems (2018)

\bibitem{he2023camouflaged}
He, C., Li, K., Zhang, Y., Tang, L., Zhang, Y., Guo, Z., Li, X.: Camouflaged
  object detection with feature decomposition and edge reconstruction. In:
  Proceedings of the IEEE/CVF Conference on Computer Vision and Pattern
  Recognition. pp. 22046--22055 (2023)

\bibitem{he2023weakly}
He, R., Dong, Q., Lin, J., Lau, R.W.: Weakly-supervised camouflaged object
  detection with scribble annotations. In: Proceedings of the AAAI Conference
  on Artificial Intelligence. vol.~37, pp. 781--789 (2023)

\bibitem{hu2023high}
Hu, X., Wang, S., Qin, X., Dai, H., Ren, W., Luo, D., Tai, Y., Shao, L.:
  High-resolution iterative feedback network for camouflaged object detection.
  In: Proceedings of the AAAI Conference on Artificial Intelligence. vol.~37,
  pp. 881--889 (2023)

\bibitem{huang2023feature}
Huang, Z., Dai, H., Xiang, T.Z., Wang, S., Chen, H.X., Qin, J., Xiong, H.:
  Feature shrinkage pyramid for camouflaged object detection with transformers.
  In: Proceedings of the IEEE/CVF Conference on Computer Vision and Pattern
  Recognition. pp. 5557--5566 (2023)

\bibitem{kim2023devil}
Kim, B., Jeong, J., Han, D., Hwang, S.J.: The devil is in the points: Weakly
  semi-supervised instance segmentation via point-guided mask representation.
  In: Proceedings of the IEEE/CVF Conference on Computer Vision and Pattern
  Recognition. pp. 11360--11370 (2023)

\bibitem{kingma2014adam}
Kingma, D.P., Ba, J.: Adam: A method for stochastic optimization. arXiv
  preprint arXiv:1412.6980  (2014)

\bibitem{kirillov2023segment}
Kirillov, A., Mintun, E., Ravi, N., Mao, H., Rolland, C., Gustafson, L., Xiao,
  T., Whitehead, S., Berg, A.C., Lo, W.Y., et~al.: Segment anything. arXiv
  preprint arXiv:2304.02643  (2023)

\bibitem{le2019anabranch}
Le, T.N., Nguyen, T.V., Nie, Z., Tran, M.T., Sugimoto, A.: Anabranch network
  for camouflaged object segmentation. Computer vision and image understanding
  \textbf{184},  45--56 (2019)

\bibitem{lee2021bbam}
Lee, J., Yi, J., Shin, C., Yoon, S.: Bbam: Bounding box attribution map for
  weakly supervised semantic and instance segmentation. In: Proceedings of the
  IEEE/CVF conference on computer vision and pattern recognition. pp.
  2643--2652 (2021)

\bibitem{liang2022tree}
Liang, Z., Wang, T., Zhang, X., Sun, J., Shen, J.: Tree energy loss: Towards
  sparsely annotated semantic segmentation. In: Proceedings of the IEEE/CVF
  Conference on Computer Vision and Pattern Recognition. pp. 16907--16916
  (2022)

\bibitem{liu2023mscaf}
Liu, Y., Li, H., Cheng, J., Chen, X.: Mscaf-net: a general framework for
  camouflaged object detection via learning multi-scale context-aware features.
  IEEE Transactions on Circuits and Systems for Video Technology  (2023)

\bibitem{liu2022convnet}
Liu, Z., Mao, H., Wu, C.Y., Feichtenhofer, C., Darrell, T., Xie, S.: A convnet
  for the 2020s. In: Proceedings of the IEEE/CVF conference on computer vision
  and pattern recognition. pp. 11976--11986 (2022)

\bibitem{lv2021simultaneously}
Lv, Y., Zhang, J., Dai, Y., Li, A., Liu, B., Barnes, N., Fan, D.P.:
  Simultaneously localize, segment and rank the camouflaged objects. In:
  Proceedings of the IEEE/CVF Conference on Computer Vision and Pattern
  Recognition. pp. 11591--11601 (2021)

\bibitem{margolin2014evaluate}
Margolin, R., Zelnik-Manor, L., Tal, A.: How to evaluate foreground maps? In:
  Proceedings of the IEEE conference on computer vision and pattern
  recognition. pp. 248--255 (2014)

\bibitem{obukhov2019gated}
Obukhov, A., Georgoulis, S., Dai, D., Van~Gool, L.: Gated crf loss for weakly
  supervised semantic image segmentation. arXiv preprint arXiv:1906.04651
  (2019)

\bibitem{pang2022zoom}
Pang, Y., Zhao, X., Xiang, T.Z., Zhang, L., Lu, H.: Zoom in and out: A
  mixed-scale triplet network for camouflaged object detection. In: Proceedings
  of the IEEE/CVF Conference on computer vision and pattern recognition. pp.
  2160--2170 (2022)

\bibitem{patrini2017making}
Patrini, G., Rozza, A., Krishna~Menon, A., Nock, R., Qu, L.: Making deep neural
  networks robust to label noise: A loss correction approach. In: Proceedings
  of the IEEE Conference on Computer Vision and Pattern Recognition (CVPR)
  (2017)

\bibitem{skurowski2018animal}
Skurowski, P., Abdulameer, H., B{\l}aszczyk, J., Depta, T., Kornacki, A.,
  Kozie{\l}, P.: Animal camouflage analysis: Chameleon database. Unpublished
  manuscript  \textbf{2}(6), ~7 (2018)

\bibitem{song2023camouflaged}
Song, Y., Li, X., Qi, L.: Camouflaged object detection with feature grafting
  and distractor aware. In: 2023 IEEE International Conference on Multimedia
  and Expo (ICME). pp. 2459--2464. IEEE (2023)

\bibitem{sun2022boundary}
Sun, Y., Wang, S., Chen, C., Xiang, T.Z.: Boundary-guided camouflaged object
  detection. arXiv preprint arXiv:2207.00794  (2022)

\bibitem{wang2022pvt}
Wang, W., Xie, E., Li, X., Fan, D.P., Song, K., Liang, D., Lu, T., Luo, P.,
  Shao, L.: Pvt v2: Improved baselines with pyramid vision transformer.
  Computational Visual Media  \textbf{8}(3),  415--424 (2022)

\bibitem{yin2022camoformer}
Yin, B., Zhang, X., Hou, Q., Sun, B.Y., Fan, D.P., Van~Gool, L.: Camoformer:
  Masked separable attention for camouflaged object detection. arXiv preprint
  arXiv:2212.06570  (2022)

\bibitem{yu2021structure}
Yu, S., Zhang, B., Xiao, J., Lim, E.G.: Structure-consistent weakly supervised
  salient object detection with local saliency coherence. In: Proceedings of
  the AAAI conference on artificial intelligence. vol.~35, pp. 3234--3242
  (2021)

\bibitem{zhang2023collaborative}
Zhang, C., Bi, H., Xiang, T.Z., Wu, R., Tong, J., Wang, X.: Collaborative
  camouflaged object detection: A large-scale dataset and benchmark. IEEE
  Transactions on Neural Networks and Learning Systems  (2023)

\bibitem{zhang2020weakly}
Zhang, J., Yu, X., Li, A., Song, P., Liu, B., Dai, Y.: Weakly-supervised
  salient object detection via scribble annotations. In: Proceedings of the
  IEEE/CVF conference on computer vision and pattern recognition. pp.
  12546--12555 (2020)

\bibitem{zhang2024cognition}
Zhang, R., Cao, Z., Yang, S., Si, L., Sun, H., Xu, L., Sun, F.:
  Cognition-driven structural prior for instance-dependent label transition
  matrix estimation. IEEE Transactions on Neural Networks and Learning Systems
  (2024)

\bibitem{zhang2023differential}
Zhang, R., Li, L., Zhang, Q., Zhang, J., Xu, L., Zhang, B., Wang, B.:
  Differential feature awareness network within antagonistic learning for
  infrared-visible object detection. IEEE Transactions on Circuits and Systems
  for Video Technology  (2023)

\bibitem{zhang2024part}
Zhang, R., Tan, J., Cao, Z., Xu, L., Liu, Y., Si, L., Sun, F.: Part-aware
  correlation networks for few-shot learning. IEEE Transactions on Multimedia
  (2024)

\bibitem{zhang2021deep}
Zhang, R., Xu, L., Yu, Z., Shi, Y., Mu, C., Xu, M.: Deep-irtarget: An automatic
  target detector in infrared imagery using dual-domain feature extraction and
  allocation. IEEE Transactions on Multimedia  \textbf{24},  1735--1749 (2021)

\bibitem{zhang2018generalized}
Zhang, Z., Sabuncu, M.: Generalized cross entropy loss for training deep neural
  networks with noisy labels. Advances in neural information processing systems
   \textbf{31} (2018)

\bibitem{zhong2022detecting}
Zhong, Y., Li, B., Tang, L., Kuang, S., Wu, S., Ding, S.: Detecting camouflaged
  object in frequency domain. In: Proceedings of the IEEE/CVF Conference on
  Computer Vision and Pattern Recognition. pp. 4504--4513 (2022)

\bibitem{zhu2022can}
Zhu, H., Li, P., Xie, H., Yan, X., Liang, D., Chen, D., Wei, M., Qin, J.: I can
  find you! boundary-guided separated attention network for camouflaged object
  detection. In: Proceedings of the AAAI Conference on Artificial Intelligence.
  vol.~36, pp. 3608--3616 (2022)

\end{thebibliography}
\end{document}